\documentclass[11pt]{article}
\usepackage[hyperref]{eacl2021}

\usepackage{times}
\usepackage{url}
\usepackage{latexsym}
\usepackage{graphicx}
\usepackage{mathtools}
\usepackage{amsmath,amsfonts,amssymb,amsthm}

\usepackage{placeins}
\usepackage{microtype}

\title{Transformer Query-Target Knowledge Discovery (TEND): Drug Discovery
from CORD-19}

\author{Leo K. Tam \\
  NVIDIA \\
  2788 San Tomas Expy \\
  Santa Clara, CA. 95051 \\
  \And
  Xiaosong Wang \\
  NVIDIA \\
  2788 San Tomas Expy \\
  Santa Clara, CA. 95051 \\
  \And
  Daguang Xu \\
  NVIDIA \\
  2788 San Tomas Expy \\
  Santa Clara, CA. 95051 \\}

\date{}

\begin{document}
\maketitle
\begin{abstract}
Previous work established skip-gram word2vec models could be used
to mine knowledge in the materials science literature for the discovery
of thermoelectrics. Recent transformer architectures have shown great
progress in language modeling and associated fine-tuned tasks, but
they have yet to be adapted for drug discovery. We present a RoBERTa
transformer-based method that extends the masked language token prediction
using query-target conditioning to treat the specificity challenge.
The transformer discovery method entails several benefits over the
word2vec method including domain-specific (antiviral) analogy performance,
negation handling, and flexible query analysis (specific) and is
demonstrated on influenza drug discovery. To stimulate COVID-19 research,
we release an influenza clinical trials and antiviral analogies dataset
used in conjunction with the COVID-19 Open Research Dataset Challenge
(CORD-19) literature dataset in the study. We examine k-shot fine-tuning
to improve the downstream analogies performance as well as to mine
analogies for model explainability. Further, the query-target analysis
is verified in a forward chaining analysis against the influenza drug
clinical trials dataset, before adapted for COVID-19 drugs (combinations
and side-effects) and on-going clinical trials. In consideration of
the present topic, we release the model, dataset, and code.\footnote{\url{https://www.kaggle.com/leotam/novel-drug-discovery-from-clinical-trials-on-dgx-2}} 
\end{abstract}

\section{Introduction}

The COVID-19 literature has experienced exponential growth and analysis
tools have arisen to digest the literature \cite{tsunami}. To mine
the literature, word embedding methods \cite{Glove, word2vec} operate
in a high-dimensional space where semantic relationships are exposed
through interrogation with metrics such as cosine similarity. Advances in transformer
attention-based architectures treat one possible weakness in word
embedding learning approaches by conditioning on contextual information
for a downstream task \cite{attentionall}. The pretrain-finetune
paradigm, whereby a language model is trained on a large corpus through
a task such as masked cloze or autoregressive variants and then trained
again (fine-tuned) for a specific application encountered success
\cite{DaiSemiSupPretrain,unifiedarch}. While word2vec implementations
focused on analogies evaluation as a downstream task indicative of
semantic learning, transformer architectures surveyed a collection
of downstream tasks such as sentiment, sentence similarity, natural
language (NL) inference, question and answering, and reading comprehension,
etc. present in the GLUE and super GLUE benchmark \cite{GLUE,superglue}.
Word2vec discovery methods \cite{latentknow} constitute a return
to analogy evaluation, though eschewing generic semantic analogies
to highlight analogies in materials science. It is hypothesized that
resolving analogies may form the basis for higher reasoning such as
advancing the limits of domain-specific research \cite{solidstatebatter,sematictext}
and measuring scholastic achievement \cite{eighthgrade}. For the
case of applications in the medical domain \cite{biobert}, transformer
pretrain-finetune performance was dependent on the quality and in-domain
nature of source and target datasets. 

 Similar to novel materials discovery, drug discovery is an intensive
and arduous process requiring trial and error. Drug discovery is the problem of allocating resources to numerous promising candidates, and thus ranking promising candidates assists the discovery process in a top-down fashion.  To date globally,
only nine influenza drugs have received full approval for use \cite{approvedantiviral}.
Where the \citet{latentknow} method uses a rigid prediction method
to mine associative analogies for undiscovered materials,
the advances in tokenization \cite{bptokenization} have relaxed exact
vocabulary registration though complicating the method. The present
work examines a query-target (QT) token conditioning method, which
extends the masked language modeling (MLM) inference. The QT method
is used to rank prediction association with clinical trials efficacy
treating a specificity problem when moving from a fixed vocabulary
to whole language tokenization. Moreover, the RoBERTa-large model trained
on the CORD-19 literature dataset (29500 provided at the time of work and now updated to over 200000 articles)
can be enhanced for drug analogies evaluation via a k-shot method
\cite{T5,gpt3}. 

\section{Related work}

Previous work has examined drug and side-effect relationships in a
bipartite graph focusing on literature in a four year range \cite{bipartite}.
While \citet{bipartite} considered 169766 PubMed abstracts across
drugs,\citet{sematictext} considered solely type II diabetes related drugs,
clustering them around heuristic expert topics.
Scoring was conducted via co-mention weighting and a five year look-back
literature mention weighting. Both previous methods incorporate expert
information in a semi-supervised method, the first through database
registration for drug - side effect pairs and the second through semantic
concept selection. The present method is an unsupervised
method excepting for physician review for the auxiliary analogies
task. Further, the method uses a majority of full texts in a narrow
focus \cite{cord19}. Results from \citet{emclinicalbert,biobert}
revealed how the detail and relevance of the dataset influences the
final result. We corroborate that domain-specific datasets improve domain-specific analogy performance.

\begin{figure*}
\centering{}\includegraphics[scale=0.46]{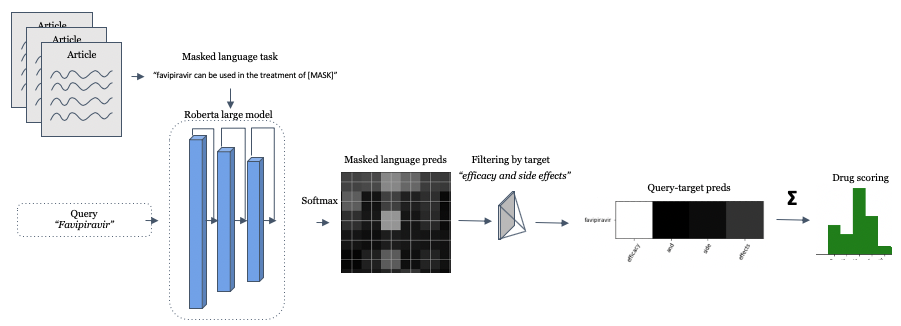}\caption{A RoBERTa-large transformer query-target method for drug discovery
reveals positive and negative associations}
\label{fig1}
\end{figure*}

\section{Methods}

The overview of our method is presented in Fig. \ref{fig1}, which
depicts both training and inference modes. During training, the MLM
task is held without modification from \citet{roberta}, namely 13.5\%
of tokens are targeted for replacement with 90\% replaced with \textless mask\textgreater ,
10\% corrupted with a random token. The MLM task is chosen as it closely replaces the function in \citet{latentknow} for predicting a target word given the context around the word.  The RoBERTa transformer method is a pure application of MLM, removing the next sentence prediction task while scaling to longer sequences with dynamic masking.  A cross-entropy loss is used for
prediction, the RoBERTa 50K byte-pair encoding tokenization is used, and hyper-parameters
are left at default settings from \citet{hugging}. During training,
the inputs are the CORD-19 dataset described in sec. \ref{subsec:Datasets}
dynamically masked ten times. The MLM training from scratch runs across 16 NVIDIA V100
GPUs in a single-node (DGX-2) configuration for 100000 steps over
approximately 36 hours. The MLM prediction is used for analogy evaluation via the structure ``A is to B as C is to \textless mask\textgreater ''
as suggested by \citet{T5}. The complete set of analogies is provided with the code.   For the word2vec implementation, the procedure
followed \citet{latentknow}, including vocabulary generation and evaluation.
The QT inference mode is discussed in Sec. \ref{subsec:Query-target-inference}. 

\subsection{Datasets\label{subsec:Datasets}}

The CORD-19 dataset is the largest machine readable full text literature
COVID-19 dataset curated by \citet{cord19}. Some statistics such as
records sourcing and license split on CORD-19 are presented in Tab. \ref{tab1}. The quality of the dataset may be attributed
to the full text access on a narrow focus that was not available in
previous medical fine-tuning studies \cite{emclinicalbert,biobert,latentknow}.
On disk as raw text, the dataset is 875 MB with 20\% of the dataset
reserved for testing during language modeling. 

\begin{table}
\begin{centering}
\caption{CORD-19 Dataset}
\label{tab1} %
\begin{tabular}{|c|c|c|c|c|}
\hline 
Sources & Records &  & License & Count\tabularnewline
\hline 
CZI & 1236 &  & Custom & 17102 \tabularnewline
\hline 
PMC & 27337 &  & Noncomm. & 2353\tabularnewline
\hline 
bioRxiv & 566 &  & Commercial & 9118\tabularnewline
\hline 
medRxiv & 361 &  & Arxiv & 927\tabularnewline
\hline 
\hline 
Total & 29500 &  &  & 29500\tabularnewline
\hline 
\end{tabular}
\par\end{centering}
\centering{}
\end{table}

\begin{figure*}
\centering{}\includegraphics[scale=0.40]{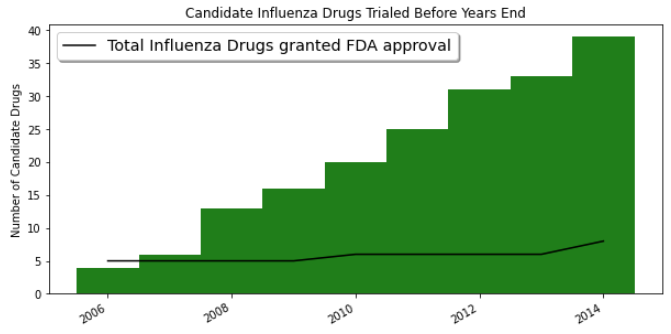}\caption{The black line represents influenza drugs receiving FDA approval.
To date, there are only eight antiviral drugs approved for influenza
strains globally with a ninth drug, remdesivir receiving emergency
approval subsequent to the analyses performed here.}
\label{fig2}
\end{figure*}

 The United States Food and Drug Administration (FDA) approved drugs
and global clinical trials data are drawn from \citet{approvedantiviral}
and \citet{trials} respectively. In Tab. \ref{tab2}, counts such
as the number of trials and drugs trialed per year's end is collected.
Namely, we use influenza as the condition, check the search term to
filter by drug treatments, and focus on years prior to 2016, when
the last antiviral drug was approved. De-duplication is performed
on trade and scientific names using the chart from \citet{approvedantiviral}.
The number of candidate and approved drugs specifically
for influenza per year is plotted on Fig. \ref{fig2}. 

 For analogy evaluation, the set of language (grammar) analogies and
drug analogies are drawn from relationships in \citet{latentknow}
and \citet{approvedantiviral} respectively. A list of the analogy
categories is presented in Tab. \ref{tab3-1}. For k-shot training,
a random set of k=5 analogies from each category is used as additional
pretraining for MLM \cite{gpt3}. 

\begin{table}
\centering{}\caption{TEND Clinical Trials Dataset}
\label{tab2} %
\begin{tabular}{|c|c|c|c|}
\hline 
Years End & Records & Drugs & Addl. Diseases\tabularnewline
\hline 
\hline 
2005 & 17 & 16 & 4\tabularnewline
\hline 
2006 & 41 & 39 & 8\tabularnewline
\hline 
2007 & 74 & 69 & 18\tabularnewline
\hline 
2008 & 112 & 107 & 30\tabularnewline
\hline 
2009 & 157 & 152 & 45\tabularnewline
\hline 
2010 & 199 & 194 & 66\tabularnewline
\hline 
2011 & 244 & 237 & 85\tabularnewline
\hline 
2012 & 275 & 268 & 102\tabularnewline
\hline 
2013 & 313 & 306 & 114\tabularnewline
\hline 
2014 & 348 & 341 & 128\tabularnewline
\hline 
2015 & 382 & 375 & 142\tabularnewline
\hline 
2016 & 411 & 371 & 157\tabularnewline
\hline 
2017 & 435 & 394 & 170\tabularnewline
\hline 
2018 & 463 & 419 & 190\tabularnewline
\hline 
2019 & 659 & 621 & 328\tabularnewline
\hline 
\end{tabular}
\end{table}

\subsection{Query-target inference\label{subsec:Query-target-inference}}

During inference, a query and target phrase are selected based on
the relationships of interest. We examine the relationship with the
RoBERTa training objective, adopting the formulation from \citet{xlnet}
for the objective as:

\begin{equation}
\underset{\theta}{\max}\log p_{\theta}(\bar{x}\vert\hat{x})\approx\underset{t}{\sum}\delta_{t}\log\frac{\exp\left(H_{\theta}^{t}(\hat{x})^{\intercal}e(x_{t})\right)}{\underset{x'}{\sum}\exp\left(H_{\theta}^{t}(\hat{x})^{\intercal}e(x')\right)}\label{eq:obj}
\end{equation}

where $\delta_{t}$ is 1 if t indicates a masked token and 0 otherwise,
$x=\left[x_{1},\cdots,x_{S}\right]$ is a text sequence, $\hat{x}$
represents a corrupted token, $\bar{x}$ represents a masked token,
$e(x)$ is the embedding of the sequence, and $H_{\theta}$ is the
RoBERTa-large architecture with parameters $\theta$ that maps a S length text sequence into
a sequence of hidden vectors. Optimizing the training objective results
in accurate MLM inference. For MLM inference \cite{bert}, the K masked
tokens in the query $q=\left[q_{1},\cdots,q_{K}\right]$ are targeted for token
prediction, i.e.

\begin{equation}
P_{k}=\frac{\exp\left(H_{\theta}^{k}(q_{k})\right)}{\underset{j}{\sum}\exp\left(H_{\theta}^{j}(q_{k})\right)}.
\end{equation}
For QT prediction, we condition the masked token targets on the query
targets $y=\left[y_{1},\cdots,y_{L}\right]$:

\begin{equation}
R\coloneqq P_{k}(q_{k}\vert q_{k}\in y)=\frac{\mathop{\underset{l}{\sum}}\exp\left(H_{\theta}^{l}(q_{k})\right)}{\underset{j}{\sum}\exp\left(H_{\theta}^{j}(q_{k})\right)}\label{eq:3}
\end{equation}
follows by independence assumption in Eqn. \ref{eq:obj} and therefore
is contained in the training objective (i.e. accurate QT prediction
is implied). When $q=y$, we observe the QT conditioning decomposes
to the MLM task prediction. The QT conditioning is more focused than
reformulating a span prediction method such as in \citet{bert} due
to rejection of extraneous tokens that would be admitted in a dot
product formulation. Once QT prediction has been formed, the analogy
MLM task may be permuted with the QT terms using ``Q is to T as Q
is to \textless mask\textgreater '' to analyze the top-k related
terms without conditioning. For rank prediction, tokens with positive
and negative associations are not intentionally mixed as they are
for visualization purposes.

\subsection{Attention visualization}

Typically transformer attention visualization examines the sequence
to sequence attention \cite{attentionall,clinicalbert}, namely plotting
the per-token attention:

\begin{equation}
A_{t}=\frac{\exp\left(H_{\theta}^{t}(x_{t})^{\intercal}e(x_{t})\right)}{\sum_{j}\exp\left(H_{\theta}^{j}(x_{t})^{\intercal}e(x_{t})\right)}.\label{eq:4}
\end{equation}
For QT visualization, the token attention per l-th query targets, namely:
\begin{equation}
R_{l}=\frac{\exp\left(H_{\theta}^{l}(q_{k})e(q_{k})\right)}{\underset{j}{\sum}\exp\left(H_{\theta}^{j}(q_{k})e(q_{k})\right)}.
\end{equation}

\begin{table*}
\centering{}%
\caption{Analogy Semantic Learning Evaluation}
\begin{tabular}{|c|c|c|}
\hline 
Category & Number & Subcategory\tabularnewline
\hline 
\hline 
drug -- inhibition & 211 & antiviral\tabularnewline
\hline 
drug -- group & 57 & antiviral\tabularnewline
\hline 
drug -- abbreviation & 57 & antiviral\tabularnewline
\hline 
drug -- approved target & 73 & antiviral\tabularnewline
\hline 
opposites & 703 & grammar\tabularnewline
\hline 
comparatives & 651 & grammar\tabularnewline
\hline 
superlatives & 651 & grammar\tabularnewline
\hline 
present participles & 4031 & grammar\tabularnewline
\hline 
past tense & 4031 & grammar\tabularnewline
\hline 
plural & 4169 & grammar\tabularnewline
\hline 
plural verbs & 993 & grammar\tabularnewline
\hline 
\end{tabular}\label{tab3-1}
\end{table*}

\begin{figure*}
\begin{centering}
\includegraphics[scale=0.49]{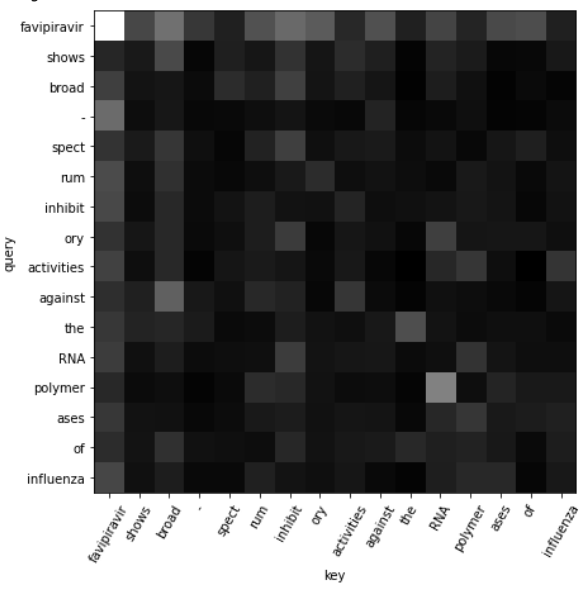}\caption{\label{fig:Attention-visualization-for}Self attention visualization (top) and query-target function (bottom
left) show associated (light) and unassociated (dark) values. A passage
from \citet{approvedantiviral} is highlighted on a per-sentence basis
using the target term ``efficacy''.}
\par\end{centering}
\includegraphics[scale=0.51]{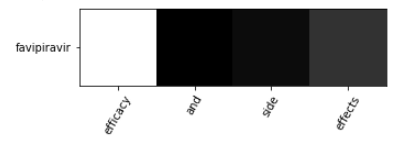}\includegraphics[scale=0.35]{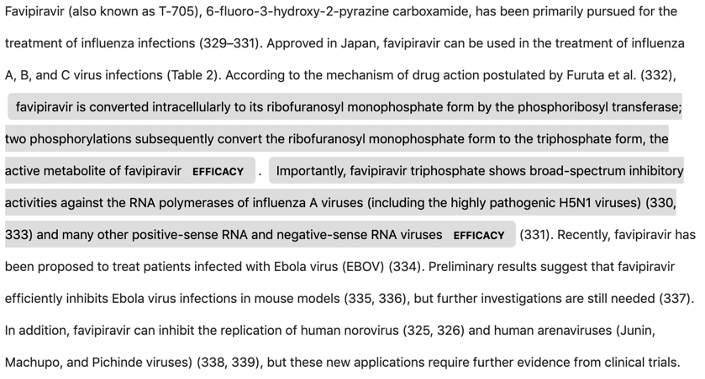}
\end{figure*}

\subsection{Forward chaining (FC) analysis}

To preserve the casual nature of time series data, the rank calculation
from Eqn. \ref{eq:3} is performed on the year-limited data in Tab.
\ref{tab3-1}. The target query is set as ``clinical trials efficacy''
and the candidate drugs are drawn from the number specified in column
2 of Tab. \ref{tab2}.  The candidate drugs are a subset of the total drugs tested as trials cover additional diseases (column 4, Tab. \ref{tab2}). 

\begin{table*}
\begin{centering}
\caption{\label{tab:Analogies-accuracy-using}Analogies accuracy using language
models trained on various corpora. {*}The word2vec cannot adequately
adapt to the phrases used in drug analogies at 600000+ vocabulary
with standard procedures.}
\begin{tabular}{|c|c|c|}
\hline 
\textbf{Model Top-5 Accuracy} & \textbf{Grammar} & \textbf{Antivirals}\tabularnewline
\hline 
\hline 
RoBERTa-large & 0.241 & 0.365\tabularnewline
\hline 
k-shot (k=5) RoBERTa-large & \textbf{0.925} & 0.538\tabularnewline
\hline 
CORD-19 RoBERTa-large & 0.727 & 0.525\tabularnewline
\hline 
k-shot (k=5) CORD-19 RoBERTa-large & 0.705 & \textbf{0.579}\tabularnewline
\hline 
 &  & \tabularnewline
\hline 
\textbf{Model Top-1 Accuracy} & \textbf{Grammar} & \textbf{Antivirals}\tabularnewline
\hline 
\hline 
RoBERTa-large & 0.082 & 0.124\tabularnewline
\hline 
k-shot (k=5) RoBERTa-large & \textbf{0.830} & 0.312\tabularnewline
\hline 
CORD-19 RoBERTa-large & 0.428 & 0.190\tabularnewline
\hline 
k-shot (k=5) CORD-19 RoBERTa-large & 0.572 & \textbf{0.396}\tabularnewline
\hline 
word2vec skip-gram & 0.504 & {*}\tabularnewline
\hline 
COVID word2vec skip-gram & 0.592 & {*}\tabularnewline
\hline 
\end{tabular}
\par\end{centering}
\end{table*}
 
\begin{figure*}
\begin{centering}
\includegraphics[scale=0.40]{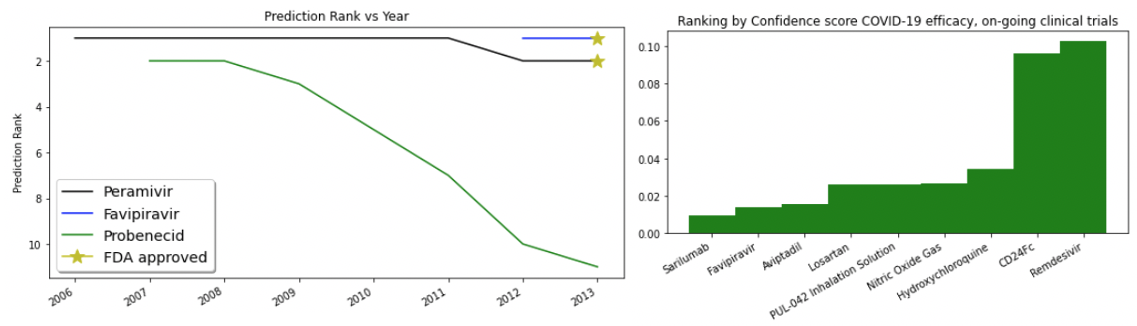}
\caption{(Left) A year-limited FC ranking analysis of influenza drugs under
clinical trials for FDA approval. Only two drugs received approval
in the period between 2005 and 2016. \label{fig:A-year-limited-(FC)}
(Right) Ranking of current clinical trials for COVID-19 drugs. (Bottom)
The permuted MLM task probes the relationship of remdesivir with clinical
trials efficacy.}
\includegraphics[scale=0.54]{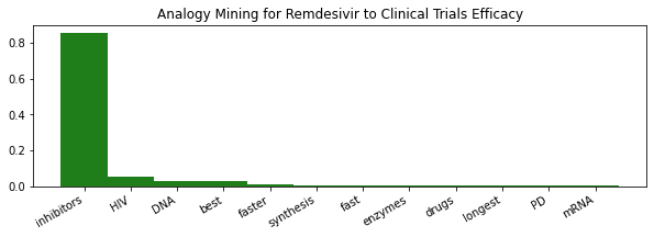}
\par\end{centering}
\end{figure*}

\section{Results}
After training for 100000 steps, the MLM task reached a perplexity
of 2.4696 on the held-out test data. The attention relationship for
self-sequence to sequence and QT is visualized in Fig. \ref{fig:Attention-visualization-for}
as per eqns. \ref{eq:3} and \ref{eq:4}. While the QT scoring may
be adapted to sentence highlighting (Fig. \ref{fig:Attention-visualization-for}),
a comparison with the span extraction or abstractive summarization
method in \citet{bert} is beyond the scope of the current work. While
negation handling (Fig. \ref{fig:Attention-visualization-for}) is
an expected result \citet{bert,superglue}, it represents an advancement
over the word2vec scoring method. Analogies evaluation is collected
in Tab. \ref{tab:Analogies-accuracy-using}. Although a comparison
on simple grammar analogies can be conducted, a simple extension cannot
be performed as the 600000+ word2vec vocabulary built from standard
procedures does not adequately capture the phrases in the drug analogies.
In the categories where RoBERTa-large can be compared to word2vec (opposites, comparatives, superlatives)
significant improvement is observed (83.0\% accuracy vs 50.4\% accuracy). The few-shot and semi-supervised
learning approaches are critical to performance, generating 23.8\%
and 18.8\% improvement in top-1 accuracy for grammar and antiviral
analogies respectively.

\begin{figure*}
\begin{centering}
\includegraphics[scale=0.27]{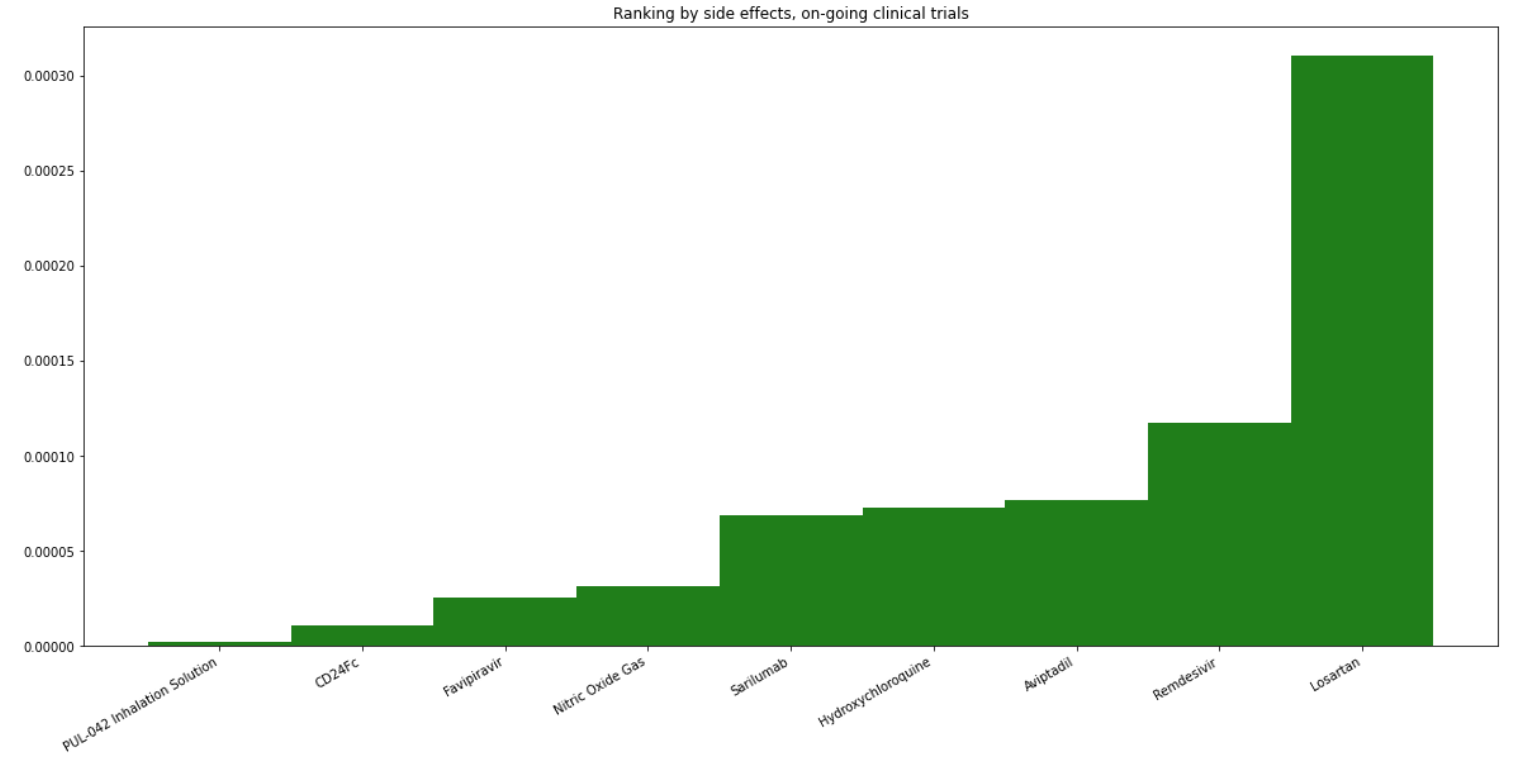}\caption{ \label{fig:fig5} (Top) Examination of side effects using the transformer query-target method. (Bottom) Drug combinations are evaluated via concatenation.}
\includegraphics[scale=0.39]{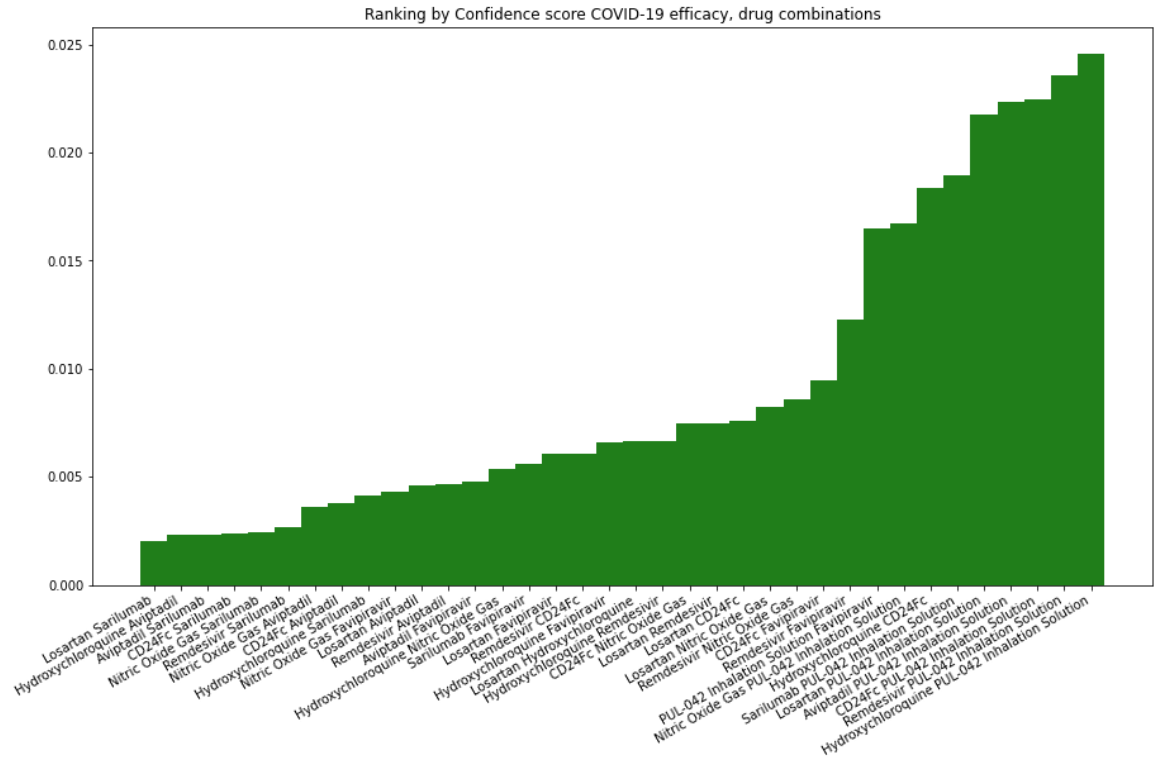}
\par\end{centering}
\end{figure*}

 While synthetic analogies can be captured to some degree by the CORD-19
RoBERTa-large model, is the model relevant for forward predictions
as in \citet{latentknow}? Fig. \ref{fig:A-year-limited-(FC)} shows
the FC analysis for the period where clinical trials data is reliably
available. Below the FC figure, a ranking of drugs under current clinical
trials is presented. Shortly after the analysis was issued ({[}anon
URL{]}), the antiviral remdesivir entered emergency FDA approval,
reflecting Fig. \ref{fig:A-year-limited-(FC)}. As a possible failure
mode, hydroxychloroquine was ranked as a distant third and was later
shown to have no correlation with positive or negative outcomes \cite{hydroxyNo}.
In Fig. \ref{fig:A-year-limited-(FC)} (bottom), the permuted MLM
task mines relationships that mirror the relationship of remdesivir
with clinical trials efficacy. Inverting the analogy mining operation
(not pictured) does not recover the QT function as predicted terms
are too generic to focus on candidate drugs. While further experiments
are collected on negative terms (side effects) and drug combinations
in Fig. \ref{fig:fig5}, a reliable method to test and verify these results
has not been collected. 

\section{Conclusion and perspectives}

A transformer QT conditioning specifies the discovery method on
a narrow literature dataset to predict clinical trials approval as
verified by FC, real-time prediction, and relationship mining. The conditioning
operation is a straight-forward calculation at inference time for the transformer
language model permissible by the independence assumption during pretraining. For language models where independence is not assumed, such as the permutation language objective \cite{xlnet}, conditioning would be performed via estimation of the posterior distribution, i.e. via a Metropolis-Hastings algorithm.  The ranking task can be used to determine
a per-sentence passage highlighting  (Fig. \ref{fig:Attention-visualization-for})
with a specific query. The scope of the QT method can
be given since $q,y\in X$ is in the set of all statements in the corpus,
and only finite sets could be generated (though by motivation this
is unwieldy). For more validation, the field of online
learning may offer independent verification through the marginal contribution to accuracy of each datum
\cite{shapley}. 

Besides the accessible resource
of clinical drug trials, other quantitative methods of determining
drug function are feasible given detailed dataset formulation. Such
methods could focus on canonical measures such as the inhibitory constant
($K_{i}$), effective dose at 95\% (ED95), or number needed to treat
(NNT). Still further are works examining protein receptor binding,
but the connection to literature machine learning methods is unclear
as well requiring specialized dataset expertise.  Due to the relatively limited number of successes for antiviral drugs, analysis suffers from sample bias. Further comparison
on the materials dataset was not possible due to unavailability of
the dataset after request.

Despite limitations, the study suggests transformer language models
are a flexible tool in mining literature.

\FloatBarrier

\bibliography{tend}
\bibliographystyle{acl_natbib}

\end{document}